# Computational Measurement of Political Positions: A Review of Text-Based Ideal Point Estimation Algorithms

Patrick Parschan[1],

and Charlott Jakob[2]

[DOCUMENT WORD COUNT: 9.863]

[ARTICLE WORD COUNT: 7763]

## ACCEPTED MANUSCRIPT

[1]Department of Media and Communication, LMU Munich

[2]Quality and Usability Lab, Technische Universität Berlin

Correspondence concerning this article should be addressed to Patrick Parschan, Department of Media and Communication, LMU Munich, Akademiestraße 7, 80799 Munich, Germany. Email: p.parschan@lmu.de

# Abstract

This article presents the first systematic review of unsupervised and semi-supervised computational text-based ideal point estimation (CT-IPE) algorithms, methods designed to infer latent political positions from textual data. These algorithms are widely used in political science, communication, computational social science, and computer science to estimate ideological preferences from parliamentary speeches, party manifestos, and social media. Over the past two decades, their development has closely followed broader NLP trends—beginning with word-frequency models and most recently turning to large language models (LLMs). While this trajectory has greatly expanded the methodological toolkit, it has also produced a fragmented field that lacks systematic comparison and clear guidance for applied use.

To address this gap, we identified 25 CT-IPE algorithms through a systematic literature review and conducted a manual content analysis of their modeling assumptions and development contexts. To compare them meaningfully, we introduce a conceptual framework that distinguishes how algorithms generate, capture, and aggregate textual variance. On this basis, we identify four methodological families—word-frequency, topic modeling, word embedding, and LLM-based approaches—and critically assess their assumptions, interpretability, scalability, and limitations.

Our review offers three contributions. First, it provides a structured synthesis of two decades of algorithm development, clarifying how diverse methods relate to one another. Second, it translates these insights into practical guidance for applied researchers, highlighting trade-offs in transparency, technical requirements, and validation strategies that shape algorithm choice. Third, it emphasizes that differences in estimation outcomes across algorithms are themselves informative, underscoring the need for systematic benchmarking.

# 1 Introduction and Research Motivation

Ideal point estimation (IPE) is a core method in the social sciences for measuring latent positions of actors, traditionally based on structured data such as roll-call votes (Poole and Rosenthal 1985). The rise of digital texts and computational power has enabled *computational text-based ideal point estimation* (CT-IPE), which infers positions directly from sources like manifestos or social media posts (Diaf 2023; Hemphill et al. 2016).

CT-IPE has produced many algorithms, from word-count methods to large language models (LLMs). While these advances broaden the methodological toolkit, they have also fragmented the field, leaving researchers with little guidance on evaluating or selecting algorithms (Benoit and Laver 2012; Bruinsma and Gemenis 2019; Egerod and Klemmensen 2020). Addressing this gap by mapping the field and supporting algorithm choice in applied research motivates our study.

Our contribution presents the first systematic review of CT-IPE algorithms (N = 25). We synthesize their development contexts and methodological logics while identifying avenues for future research. We address two research questions (RQs):

RQ1: What is the development context under which CT-IPE algorithms have been developed?

RQ2: How do algorithms model variance to estimate latent political ideal point constructs?

In RQ1, we examine the contexts in which algorithms were developed, such as political systems and textual genres. In RQ2, we group CT-IPE algorithms into four methodological types, building on prior research and extending it inductively by analyzing how each algorithm models textual variance.

The paper proceeds as follows: Sections 2 and 3 situate CT-IPE within the broader literature and present our research questions. Section 4 outlines data collection and annotation. Section 5 reports results, moving from development context (RQ1) to our typology (RQ2) and practical guidance for algorithm choice. Sections 6 and 7 conclude with limitations, implications, and future directions.

## 2 Ideal Points and Ideal Point Constructs

IPE is the "generation of meaningful measures" for the "use in studies requiring continuous measures of latent preferences" (Carroll 2023). We borrow from Bafumi et al. (2005) to lay the definitional foundation for our review. They define ideal points as "preferences or capacities within a spatial framework." This spatial framework is one-dimensional (a line) in its most straightforward shape but can have higher dimensions. Within it, any political actor's latent "disposition/preference can be depicted by a point on this line—the person's [or political actor's] ideal point (Bafumi et al. 2005, 171).

We refer to these latent preferences as *ideal point constructs* and connect to Laver (2001, 68), who calls ideal points "unobservable constructs buried deep in the brain." We want to emphasize their nature as unobservable social variables: abstract positions taken toward political issues (e.g., gun control) or ideological dimensions (e.g., left-right).

In this sense, positioning involves more than just preference. As Van Langenhove (2020, 5) puts it, positions "can be described as clusters of beliefs that people have with respect to the rights and duties to act in certain ways. "Positioning" refers to the processes of assigning, appropriating, or rejecting positions."

To measure such theoretical positions and positioning, CT-IPE builds on other approaches to IPE. It extends manual text-based IPE methods like the *Manifesto Project* (Lehmann et al. 2024), eliminating the need for extensive human labor. Additionally, CT-IPE builds on earlier computational IPE methods by using unstructured text data instead of the structured data typically employed in approaches like the *NOMINATE* algorithm, which relies on roll-call votes (Poole and Rosenthal 1985).

# 3 Related Work and Research Questions

Several prior reviews informed ours. Mair (2001) and Volkens (2007) provide early overviews of human-based IPE and some CT-IPE methods. Although not computational in focus, they shaped our expectations around what ideal point constructs CT-IPE algorithms estimate.

More recent works like Egerod and Klemmensen (2020) or Hjorth et al. (2015) compare a limited number of CT-IPE algorithms or examine related strands of literature like *political viewpoint identification* (Doan and Gulla 2022) and *position scaling* (Abercrombie and Batista-Navarro 2020).

While all different in scope, these studies helped identify key dimensions of comparison, such as actor type, text genre, and validation strategy, which we refer to as variables of the *development context*. We define *development context* as a set of substantive, theoretical, and empirical parameters that shape the design of a CT-IPE algorithm. These elements determine the algorithm's scope, assumptions, and applicability and are central to understanding how methodological choices align with the goals of the algorithm's design. As our first RQ, we therefore ask:

**RQ1***: What is the development context under which CT-IPE algorithms have been developed?*

While the development context shapes methodological decisions, CT-IPE is fundamentally a measurement problem: latent constructs are not directly observable, and researchers infer them from systematic variance in text data. Algorithms estimate ideal points by modeling this textual variance. Therefore, we are not only interested in the development context and its surrounding constraints, but also in how the measurement problem itself has been approached.

Existing literature highlights the central role of variance in this regard. Van der Brug (2001, 126) talks about "explained variance" in estimating the ideal points of parties in the Netherlands. McDonald (2013, 119) distinguishes irrelevant variance from "the proportion of variance that comes from the actual underlying party positions." Budge and Meyer (2013) contrast temporal variance in manifesto data with more static expert judgments, underscoring that finding relevant variance is central to IPE.

Given the general importance of variance for measuring latent constructs as manifest variables in the social sciences, it is unsurprising that this is the common denominator around which all CT-IPE algorithms can be grouped.

We build on this by offering a unifying framework for understanding CT-IPE algorithms: all of them (1) generate, (2) capture, and (3) aggregate textual variance to produce ideal point estimates. This framework guides our RQ2:

***RQ2***: *How do algorithms model variance to estimate latent political ideal point constructs?*

    ***RQ2.1***: *How do algorithms transform text into numerical data to generate variance?*

    ***RQ2.2***: *How do algorithms capture variance relevant to latent ideal point constructs?*

    ***RQ2.3***: *How do algorithms aggregate relevant variance into an ideal point estimate?*

Prior research has emphasized the importance of accurately capturing variance and the problem of "improper variance estimates" (Bafumi et al. 2005, 171). Lowe (2008, 359), in a widely cited paper, investigates how the "policy position variance" of the *Wordscores* algorithm might be distorted, yet again underscoring how important variance is for CT-IPE. Also, a meta-methodological article by Benoit and Laver (2012, 205) asks whether some methods "capture significantly more of the variance" relevant for measuring points in a spatial political model. Early work, notably Anthony Downs, also emphasizes variance: he argues that the "factors in our model explain how wide ideological variance can develop" (Downs 1957, 100).

Given this deduction from existing literature, variance as a conceptual measurement logic offers a unifying framework for understanding CT-IPE algorithms at a high level of abstraction, one necessary for synthesizing a diverse field. Our two research questions (RQs) are designed to characterize the algorithms we identify. They link the development context in which these algorithms emerge with the fundamental measurement decisions that shape how CT-IPE is approached. The two RQs will guide the development of our review scope and content analysis codebook.

# 4 Systematic Review

Our review builds on two pillars: (1) explicit inclusion criteria that define the scope of CT-IPE algorithms considered (see section 4.1), and (2) a transparent multi-stage sampling and annotation process (see sections 4.2 and 4.3). The criteria establish the boundaries of what counts as a CT-IPE algorithm, while the process ensures that these criteria are applied consistently and replicably. Together, they form the foundation of our algorithm selection.

## 4.1 Scope of the Review

We conducted a systematic literature review guided by four inclusion criteria to identify and compare the diverse approaches.

*Criterion (1): The algorithm aims to estimate scalar, continuous numerical values (i.e., ideal points) for political actors.*

We included only peer-reviewed algorithms explicitly developed for CT-IPE, excluding those with merely theoretical potential. Evaluating such possible applications would require in-depth assessments beyond the scope of this review.

*Criterion (2): The algorithm operates solely on raw text data.*

The core input must be text authored by a political actor—e.g., a social media post or a manifesto—and an associated actor identifier. The algorithm must rely solely on the text itself and its formal metadata (e.g., speaker, party, session). Approaches that require additional external data (e.g., constituency or voting behavior) fall outside the scope of this review.

*Criterion (3): The algorithm is unsupervised or semi-supervised.*

We understand semi-supervised learning as "unsupervised learning guided by constraints," for instance, by selecting keywords or reference documents (Chapelle et al. 2006, 2). We exclude supervised approaches, which rely on extensive human labeling of many individual data rows.

*Criterion (4): The algorithm is a formal methodological contribution.*

We included papers that either formally (a) propose a new CT-IPE algorithm (often signaled by a distinct name) or (b) modify an existing CT-IPE algorithm. While development is often iterative, we aim to map core contributions, which link each algorithm to a central paper.

These criteria give clear inclusion boundaries to capture the most relevant developments in the field. They ensure a transparent and replicable selection process while focusing our review on established CT-IPE contributions.

## 4.2 Sampling Process

### 4.2.1 Database Search Methodology

Guided by our RQs and an initial exploration (see Appendix), we performed database searches across four sources: Web of Science, EBSCOhost, arXiv, and the ACL Anthology. We used a Boolean search string composed of 214 keywords to capture relevant, peer-reviewed literature (see online Appendix 1 for the whole query).[1]

We limited our search to 35 years from January 1, 1990, to January 23, 2025. We performed all database searches in the last week of January 2025. An exploratory phase of the literature review informed this timeframe, during which we found no relevant records before 1990 (see also Figure 2). Where available, we applied filters for database sub-indices, subject areas, or research fields to narrow our search.[2]

---

[1] For the ACL anthology, no web interface exists. We downloaded the database file here: https://aclanthology.org/ We then used the python package eldar to query the database file with the same boolean search string we use for the other databases (Chenebaux 2022). The code for this can be found in online Appendix 6.

The initial queries yielded 25,411 records across the four databases. We removed duplicates using the `dedup` function from ASReview, an increasingly prominent machine-assisted literature review tool with various functionalities (Van De Schoot et al. 2021). We then applied heuristic keyword filtering to remove further irrelevant entries (see online Appendix 6 for the filtering script). The PRISMA chart in Figure 1 and Table A1 in the Appendix provide a detailed overview of the sampling process.

---

[2] A full list of the searched indices, subjects, and research areas can be found in online Appendix 1.

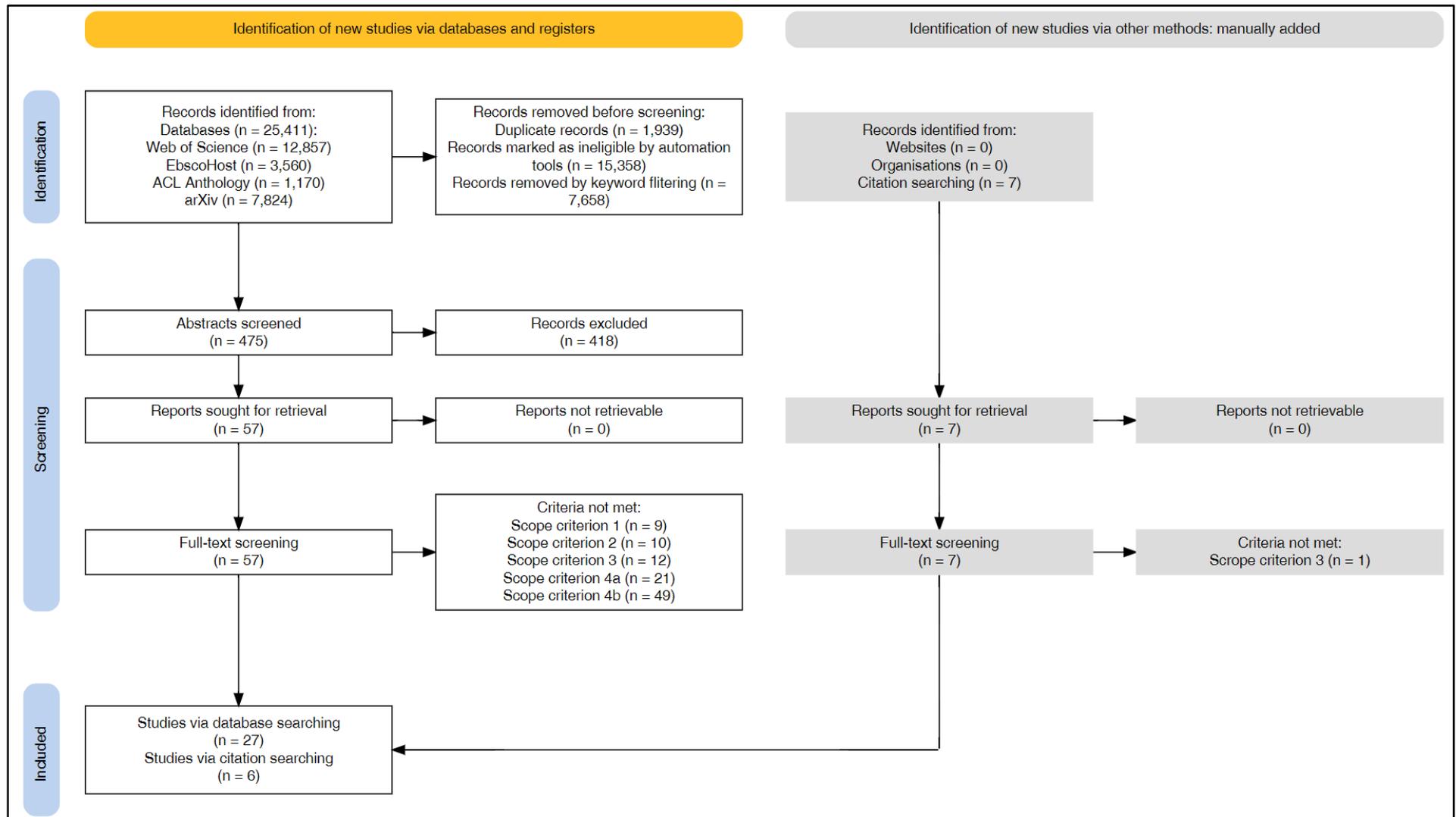

Figure 1: PRISMA chart. Note: Criteria 4a and 4b are not mutually exclusive, so the total number of criteria-triggered exclusions is higher than the number of excluded papers in the "Criteria not met" box. In total, 31 papers were excluded at this stage, leaving 33 of the original 64 for inclusion. For details, see Online Appendix 3.

## 4.2.2 Abstract Screening

We loaded the resulting 15,833 records[3] into ASReview and used its active learning functionality to screen titles and abstracts. In cases of uncertainty, we opted for inclusion to allow for more robust decisions during the full-text screening phase.

Active learning dynamically reorders the dataset after each labeled record. With every new annotation, the tool trains a new model, re-ranks all remaining unlabeled records by predicted relevance, and presents the currently most relevant instance to the researcher for labeling[4] (Van De Schoot et al. 2021). This enables efficient screening without reviewing the entire corpus, a key advantage for large-scale reviews. Finding all relevant papers may be unlikely, but one will still get a large sample of relevant papers. Screening tens of thousands of papers is infeasible for most review projects, and even in smaller samples, human fatigue can lead to missed papers and errors in judgment.

The question then is, "how to decide when to stop reviewing so that 1) most relevant papers have been retrieved, 2) not too much review effort is wasted" (Yu and Menzies 2019, 12). Stopping after 50 consecutive non-relevant records is usually "too early and results in a low recall" (Yu and Menzies 2019, 13).

Therefore, we adopted a conservative strategy: each author independently screened abstracts until they encountered 100 consecutive entries marked as non-relevant. This led Author 1 to screen 320 records and Author 2 to screen 453. In total, we identified 57 abstracts for full-text screening.[5]

---

[3] A dataset of these 15,833 records can be found in online Appendix 10.

[4] Our models are based on term frequency–inverse document frequency (Tf-idf) featurization combined with a Naive-Bayes classifier. This setup has "high performance" and "low computation time." (Van De Schoot et al. 2021, 128) The annotations of both authors can be found in online Appendix 2, 8, and 9. Training data for the model is still needed before the active learning cycle starts, just as would be with any other machine learning model. For both authors, as prior knowledge (i.e. training data), we used 14 papers as "include" and five as "not include". The 19 papers were found in the exploration (see Appendix) and both authors agreed regarding their fit or non-fit to the review. See the R script in online Appendix 6.

[5] Excel files and ASReview project files for the abstract screening can be found in online Appendix 2, 8, and 9. See also the textual Appendix for a more detailed documentation of this step.

### 4.2.3 Full-Text Screening

We retrieved and read all 57 full texts. We assessed each against the four inclusion criteria from section 4.1. Both authors independently coded each text according to these criteria.[6] In addition, we used backward citation tracking and identified seven more potentially relevant papers, which we screened and annotated in the same way. This brought the total to 64 full-texts.

We achieved strong inter-coder agreement on these 64 documents: a .86 simple agreement and a Krippendorff's alpha of .72. We resolved all nine disagreements through discussion. Ultimately, 33 papers met the criteria and were included in our final sample for manual content analysis. Data for our annotations of this step can be found in online Appendix 3.

## 4.3 Codebook for Standardized Content Analysis

We conducted a qualitative content analysis using a standardized annotation codebook to answer our research questions. Our process follows the STAMP guidelines, which recommend "copying the text passages that were decisive for the coding" (Rogge et al. 2024, 10). We identified and annotated the passage in the full text for each variable that best matched its definition and annotation instructions. We provide an overview of all variables in the online Appendix 4.

Each author independently annotated all 33 papers using a set of 18 variables. These were designed to answer our two RQs. The 15 RQ1-focused variables capture the development context of algorithms, such as country, actor type, and ideal point construct. RQ2-focused variables capture methodological choices—how algorithms generate, capture, and aggregate variance. After independently annotating all variables and understanding the algorithms even deeper, we again excluded eight papers, leading to a final sample of 25 papers.

---

[6] See online Appendix 3.

While the 15 RQ1 variables are more numerous and relatively straightforward, the three RQ2 variables are fewer but more complex. Below, we briefly illustrate our rationale using selected examples (see Table 1). For the complete codebook, including variable names, descriptions, annotation instructions, and example annotations, we refer the reader to online Appendix 4.

To address RQ1, we focused on the context under which algorithms were developed, annotating 15 variables for every paper. For instance, we annotated the countries or regions, typically drawn from direct mentions (e.g., "United States"). The second example is the political actors (e.g., "members of Congress") for whom a paper performs CT-IPE.

Another example of a more complex variable related to RQ1 is cross-data transferability. This refers to whether authors claim that their algorithm can generalize across different data contexts beyond the one for which it was originally developed. For instance, this could involve applying an algorithm trained on political manifestos to social media data.

To address RQ2, we annotated three methodological variables (see Table 1) aligned with our framework. To answer RQ2.1, we annotated how algorithms generate variance through featurization, such as counting words. RQ2.2 looks into how algorithms capture variance relevant to point estimation. In this step, algorithms often identify reference (pivot) texts or keywords that represent the ends of an axis. RQ2.3 asks how variance is aggregated into scalar point estimates. Here, we found semi-supervised methods like cosine similarity or unsupervised methods like principal component analysis (PCA).

| countries_or_regions (RQ1) | actor_scaled (RQ1) | cross_construct_transferability (RQ1) | generate_variance (RQ2.1) | capture_variance (RQ2.2) | Aggregate_variance (RQ2.3) |
|---|---|---|---|---|---|
| "U.S." | "members of Congress (MoCs)" | "Like Wordscores, #Polar-Hashtag scores are language- and institution-independent and can be used to estimate the ideology of any speaker based on the text produced on Twitter." | "word frequency" | "the choice of focal words" | "If a given post is proximate to the 'hate' word vector, we thus classify it as more hateful on average. The typical distance measure is the cosine similarity measure" |
| "United States" | "presidential candidates and members of Congress" | "Our approach can be applied to media other than newspapers if news reports are converted into text data." | "word embeddings" | "We name these two most dissimilar texts pivot texts and assign an initial position score of 1 to one of them and −1 to the other" | "To generate rankings from our data, we apply principal components analysis (PCA) to the parties' opinions and project them on the first few principal components." |

Table 1: Example annotations.

# 5 Results

## 5.1 RQ1: Development Context

### 5.1.1 Social Science Theory, Ideal Point Constructs, and Nomenclature

Many reviewed papers engage with social science theories underpinning their algorithms. For example, some papers explicitly reference spatial political models (e.g., Laver et al. 2003; Slapin and Proksch 2008; Nanni et al. 2022), while others use theoretical ideas around framing (Hemphill et al. 2016; Vafa et al. 2020), gatekeeping (Kaneko et al. 2021), cleavages (Gentzkow et al. 2019), political communication (Lo et al. 2014; Temporão et al. 2018), party-conflict (Lauderdale and Herzog 2016), and polarization (Burnham 2024).

Due to word constraints, we limit our further discussion of RQ1 to the main findings without citing individual papers. We refer readers to online Appendix 5 for detailed annotations of each paper.

Connected to the theoretical diversity is the fact that the constructs employed in the papers of our review extend beyond the left-right construct, encompassing the following dimensions: economic, societal, foreign policy, external relations, freedom and democracy, conservative and liberal ideologies, authoritarian vs. libertarian, European integration, political hostility, government–opposition, cultural axis, gun control, abortion rights, anti-subsidy vs. pro-subsidy, and even more niche constructs like 'tea-partiness'.

The varied nomenclature across the literature again mirrors this diversity in constructs. Some authors refer to their methods as scaling, estimation, measuring, or even extraction. We propose adopting *ideal point estimation* as a consistent term, as it best captures the statistical methods and conceptual foundations common to these studies and aligns with the long tradition of non-textual IPE in political science.

**5.1.2 Political Actors, Countries, and Data**

Our review indicates that all algorithms were developed with political actors as their primary focus. Algorithm development concentrates on politicians, from senators and candidates to influential social media figures. However, political parties are equally important, while only some algorithms are developed with non-elite voters in mind. The number of entities analyzed is highly variable: some studies focus on as few as two entities (for example, in inter-party comparisons), whereas others analyze tens of thousands, particularly in social media-based research.

The geographic focus is mainly on Western democracies and their languages, but also includes countries from Asia, such as Japan. The data spans a broad temporal range from 1854 to 2023, focusing on the 20th century. Data sources vary, but concentrate on party manifestos, speeches, and social media posts as the data sources. The number of data points ranges from a few dozen documents (more typical for manifestos) to millions (as seen with social media posts), which poses unique challenges for modeling decisions.

**5.1.3 Validation, Transferability, and Open Science Practices**

Every paper in our review performs some validation against external gold standards like more established algorithms or NOMINATE scores. In the lingo of our review, validation is concerned with ensuring that the algorithm models variance correctly. Many papers use face validity and leverage established human-annotated data projects, such as the Manifesto Project and expert surveys, to validate their algorithms. Others conduct their own manual annotations to ensure validity. Almost always, the papers include an illustrative study in which the algorithm is applied to showcase the validity of the new measurements. In section 5.3.3, we go into more detail about how algorithms validate their approaches and what practical implications this has.

Connecting to validity, authors discuss whether their approaches can be applied to different settings, such as data types or ideal point constructs. While some reviewed papers go into these aspects, they do not devote much space to explicit claims about transferability, suggestions for further research, or systematic evaluations across domains.

At the same time, open science practices are also strongly evidenced. There is a robust but not universal culture of sharing, with most algorithms making their code (see Figure 8) or data publicly available. However, especially for older papers, the sharing URLs are sometimes dead.

To summarize RQ1 results, we find that CT-IPE algorithms develop across diverse contexts. They target ideal point constructs beyond left-right ideology and apply to various actor types, countries, and text genres. Validation practices are well established. Papers discuss transferability, and open science practices, especially code and data sharing, are standard. This diversity underscores the need for deeper methodological integration of different algorithmic approaches.

Together, these findings provide a structured account of the contexts in which CT-IPE algorithms were developed—across countries, actors, genres, validation strategies, and transferability. This forms the first side of our analysis and sets the stage for the methodological perspective in RQ2.

## 5.2 RQ2: Four Typical Approaches

While RQ1 examined the development context, RQ2 focuses on their methodological structure. We group CT-IPE algorithms into four approaches (Table 2), each reflecting broader NLP trends (see Figure 2) and differing in how they generate, capture, and aggregate textual variance:

(I)    Word frequency-based

(II)    Topic modeling-based

(III)    Word embedding-based

(IV)    LLM-based

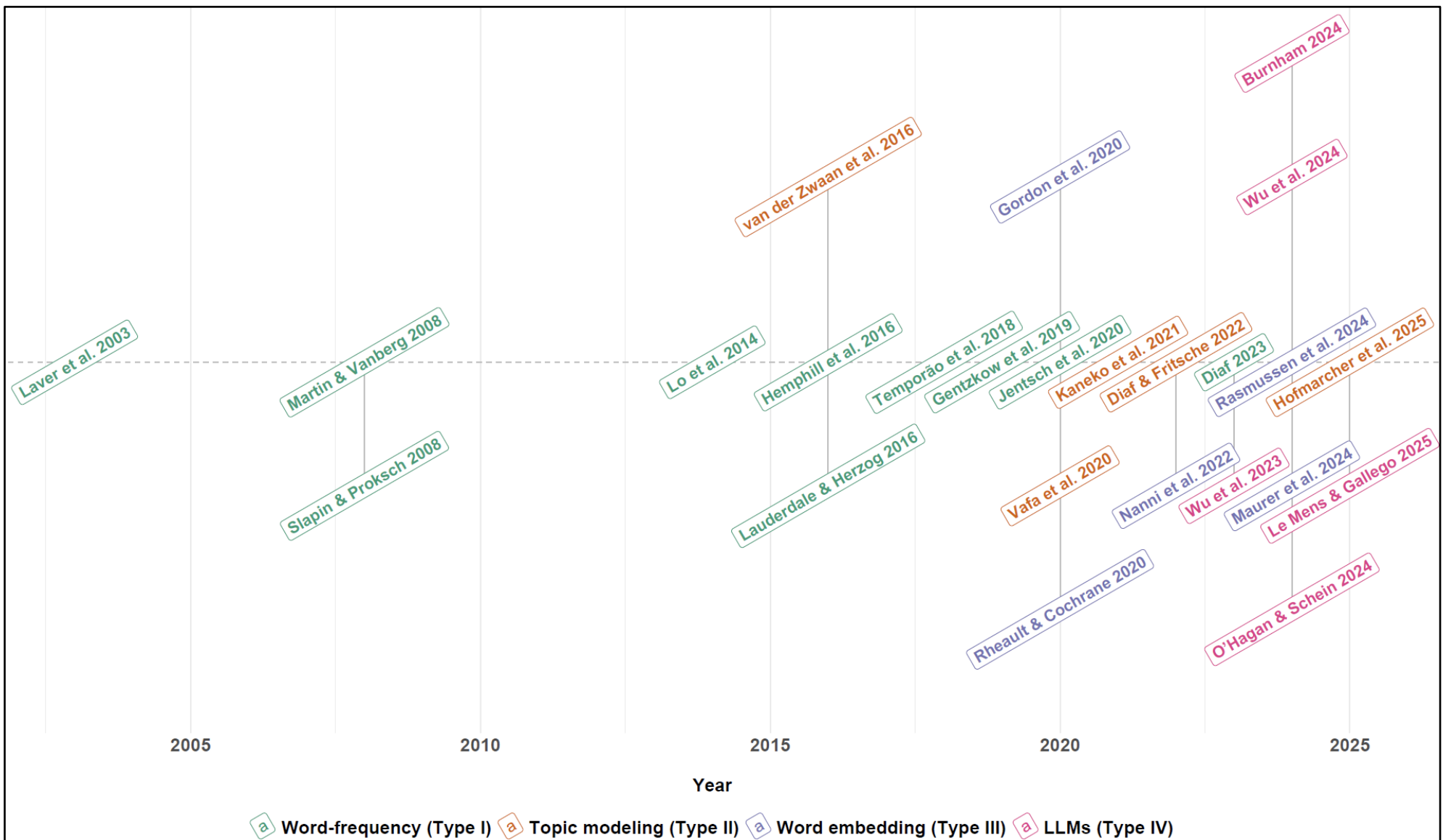

Figure 2: Development timeline of all 25 CT-IPE algorithms.

Combining our three variance-related variables forms a strong basis for the typology. The steps of generating, capturing, and aggregating variance are interdependent. For example, in Type I algorithms, using word frequencies influences how variance is captured and aggregated subsequently. Similarly, a Type II algorithm may use word embeddings to generate topics but is not Type III (word-embedding-based) if conceptual topics are the input to point estimation. This shows that categorization depends on the full algorithmic setup. These dependencies lead to clear-cut types, leaving no fringe cases. The typology emerged inductively during our annotation process. We did not begin with fixed types but observed them as we annotated the methodological variables.

We emphasize that the following high-level frameworks are conceptual tools, designed to bring analytical coherence to a diverse field, not to represent the full complexity of single algorithms. They serve to reflect on how groups of CT-IPE algorithms engage with the IPE measurement problem.

In what follows, we outline how each type works, using our generate–capture–aggregate framework to describe the internal logic of each group. We focus on comparative insights rather than technical detail; full descriptions and annotated variables are provided in the cited papers and in online Appendix 5.

| Paper | Algorithm Name | Comment | Type | Reviewed |
|---|---|---|---|---|
| Laver et al. (2003) | Wordscores | - | Type I | yes |
| Martin & Vanberg (2008) | - | Wordscores extension | Type I | yes |
| Slapin & Proksch (2008) | Wordfish | - | Type I | yes |
| Lo et al. (2014) | - | Wordfish extension | Type I | yes |
| Hemphill et al. (2016) | #Polar Scores | Wordfish extension | Type I | yes |
| Lauderdale & Herzog (2016) | Wordshoal | Wordfish extension | Type I | yes |
| Temporão et al. (2018) | Dynamic Lexicon Approach | Wordfish extension | Type I | yes |
| Gentzkow et al. (2019) | - | - | Type I | yes |
| Jentsch et al. (2020) | Time-dependent Poisson reduced rank models | Wordfish extension | Type I | yes |
| Diaf (2023) | CommunityFish | Wordfish extension | Type I | yes |
| van der Zwaan et al. (2016) | Cross-Perspective Topic Modeling (CPT) | CPT invented earlier | Type II | yes |
| Vafa et al. (2020) | Text-Based Ideal Point Model (TBIP) | - | Type II | yes |
| Kaneko et al., (2021) | - | - | Type II | yes |
| Diaf & Fritsche (2022) | TopicShoal | - | Type II | yes |
| Hofmarcher et al. (2025) | Time-Varying Text-Based Ideal Point Model (TV-TBIP) | Based on Vafa et al. (2020) and Gentzkow et al. (2019) | Type II | yes |
| Gordon et al. (2020) | - | - | Type III | yes |
| Rheault & Cochrane (2020) | Party Embeddings | - | Type III | yes |
| Nanni et al. (2022) | SemScale, SemScore | - | Type III | yes |
| H. R. Rasmussen et al. (2024) | Super-Unsupervised Classification (SU) | | Type III | yes |
| Maurer et al. (2024) | SBERTHashtag | - | Type III | yes |
| Wu et al. (2023) | Language model pairwise comparison (LaMP) | - | Type IV | no |
| Wu et al. (2024) | Concept-guided chain-of-thought (CGCoT) | Based on LaMP | Type IV | yes |
| O'Hagan & Schein (2024) | - | - | Type IV | under review |
| Burnham (2024) | Semantic Scaling | - | Type IV | no |
| Le Mens & Gallego (2025) | Asking and Averaging | | Type IV | yes |

Table 2: Overview of all reviewed CT-IPE algorithms. Type I: Word frequency-based algorithms; Type II: Topic modeling-based algorithms; Type III: Word embedding-based algorithms; Type IV: LLM-based algorithms. We relaxed the need for a peer-review status for LLM-based algorithms since these have often not yet had the time to undergo the whole process.

### 5.2.1 Type I: Word Frequency-based Algorithms

Type I algorithms represent text through raw word frequencies and apply statistical models to uncover patterns of variance that reflect underlying political positions. This approach underlies some of the earliest and most widely used CT-IPE methods, including Wordscores (Laver et al. 2003). *Wordscores* is semi-supervised and captures variance via reference texts to estimate ideal points. Although sometimes labeled as a supervised method, *Wordscores* does not rely on extensive human annotations, but rather a limited number of anchor texts, placing it at the semi-supervised end of the spectrum. It has been widely adopted and discussed (Lowe 2008; Bruinsma and Gemenis 2019). The general logic of Type I follows a three-step process:

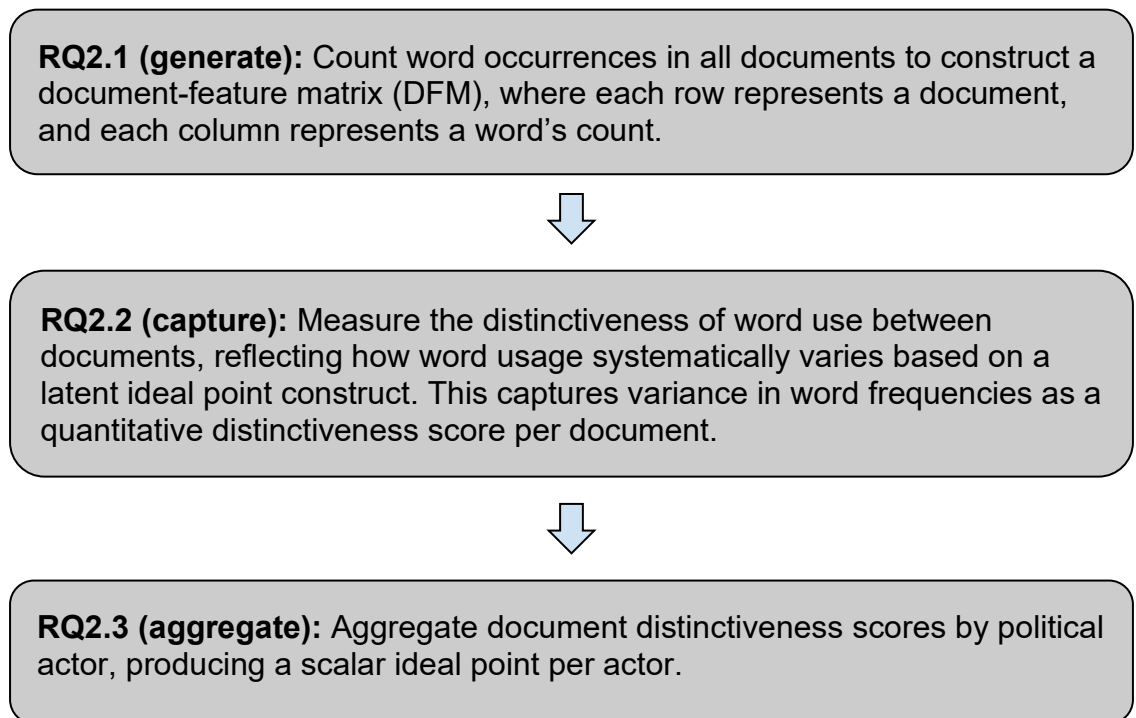

**RQ2.1 (generate):** Count word occurrences in all documents to construct a document-feature matrix (DFM), where each row represents a document, and each column represents a word's count.

**RQ2.2 (capture):** Measure the distinctiveness of word use between documents, reflecting how word usage systematically varies based on a latent ideal point construct. This captures variance in word frequencies as a quantitative distinctiveness score per document.

**RQ2.3 (aggregate):** Aggregate document distinctiveness scores by political actor, producing a scalar ideal point per actor.

Figure 3: Type I framework.

Another example is *Wordfish* (Slapin and Proksch, 2008), a fully unsupervised Type I algorithm. It generates variance as word counts drawn from a Poisson distribution. It captures variance through word-specific discrimination parameters, linking word frequency differences to latent political positions. It then aggregates variance by using maximum likelihood estimation (MLE), where the ideal point for each document or political actor is one variable in the equation (for an evaluation and discussion of Wordfish, see Hjorth et al. (2015)). Variants include *Wordshoal* (Lauderdale and Herzog 2016) and *CommunityFish* (Diaf 2023), adding new aggregation methods to the *Wordfish* algorithm using factor analysis and clustering. Jentsch et al. (2020) also extend *Wordfish* and propose a time-varying extension.

Lowe (2016, 2) summarized that the core idea behind Type I algorithms is to extract "patterns of relative emphasis" from word counts. That is, algorithms generate variance by counting words per document and then capture variance by looking for differences in word counts. Lastly, they aggregate variational patterns across each political actor's documents. The focus on word count differences between documents makes this type methodologically distinct from later methods that rely on semantic or neural representations of text.

### 5.2.2 Type II: Topic Modeling-based Algorithms

Unlike Type I algorithms, which use raw word frequencies as features, topic modeling introduces an intermediate layer: it represents documents as distributions over topics to generate variance. This added level of abstraction allows Type II algorithms to treat ideal points as functions of topic alignments, which Grimmer et al. (2022, 236) describe as "a mixture across categories."

An illustrative example is the *Text-Based Ideal Point (TBIP)* algorithm by Vafa et al. (2020). Using Poisson factorization, variance is generated by representing each document through vectors of topic intensities. It captures variance by combining these neutral topics with ideological adjustment vectors, which shift word use depending on an author's latent ideal point. It then uses Variational Inference (VI) to aggregate variance by jointly inferring each author's ideal point together with the topics during model estimation. Type II follows a general framework:

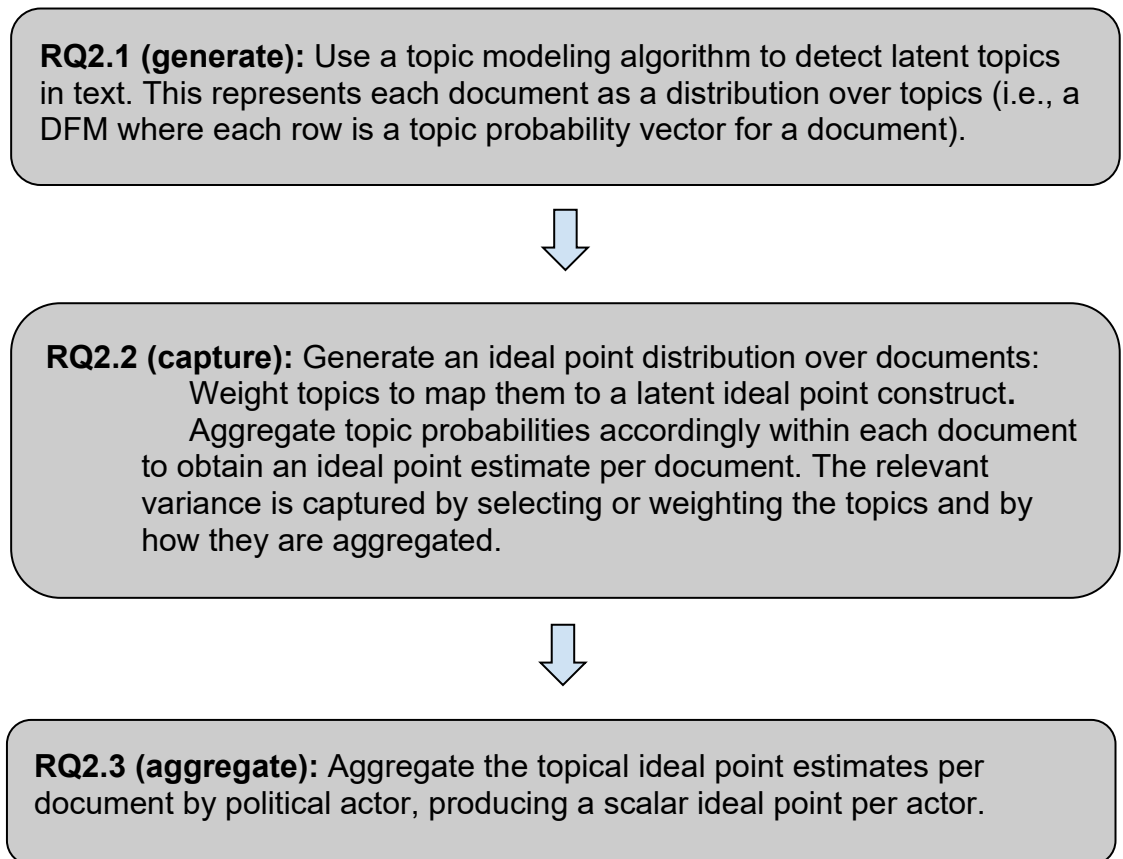

Figure 4: Type II framework.

Van der Zwaan et al. (2016) refine and validate *the Cross-Perspective Topic Model (CPT)* for ideal point estimation of parties. The algorithm generates variance by training two Latent Dirichlet Allocation (LDA) topic models separately: one on nouns (topics) and one on adjectives, verbs, and adverbs (opinions). It captures variance by associating each topic with different sets of the so learned opinions, so that parties can express distinct viewpoints on the same topics. It aggregates variance by inferring party-level opinion profiles using Gibbs sampling, a Markov Chain Monte Carlo (MCMC) algorithm.

Hofmarcher et al. (2025) propose *TV-TBIP*, building on Vafa et al. (2020) and Gentzkow (2019). This time-varying version of the *TBIP* model incorporates the idea that ideal points change over time. It thus addresses temporal dynamics in political positioning, including time as formal metadata (see scope criterion (2)) in the estimation.

Type II algorithms estimate ideal points using topic modeling. This abstraction allows for flexible mappings to political space. While most models remain grounded in probabilistic topic modeling, some, like *TopicShoal* (Diaf and Fritsche 2022), use word embeddings to generate topic representations. Overall, Type II methods reflect a distinct modeling paradigm emphasizing thematic structure as relevant for representing ideal points.

### 5.2.3 Type III: Word Embedding-based Algorithms

Word embeddings are a "special type of distributed word representation that are constructed by leveraging neural networks" (Pilehvar 2022, 25). Learned from large corpora, they map words, sentences, or entire documents to high-dimensional numerical vectors, where vector distances encode semantic similarity. CT-IPE algorithms in Type III estimate ideal points by capturing and aggregating variance in these high-dimensional numerical vectors.

The most intuitive algorithm of this type is the *Party Embeddings* algorithm by Rheault and Cochrane (2020), which is fully unsupervised. The authors generate variance by training word embeddings on parliamentary speeches, augmented with metadata such as party affiliation. The algorithm captures variance implicitly, as the authors assume that the first principal components of the embedding space reflect ideal point constructs like the left-right dimension. It aggregates variance by projecting party embeddings onto these components, producing ideal point estimates for each party on two unsupervised axes.

A key difference between this type and types I and II lies in how variance is generated: whereas Type I uses raw word counts and Type II abstracts them into probabilistic topic distributions, Type III directly embeds words, sentences, or documents into dense semantic vector spaces using neural models. Because of this, algorithms in this group can capture variance at the word, sentence, or document level.

Hebbelstrup Rye Rasmussen et al. (2024) propose the *Super-unsupervised Classification* algorithm. It generates variance from word embeddings trained on political texts. It captures variance by selecting keywords that are relevant to the ideal point construct. It aggregates variance by computing similarity scores between keyword embedding vectors and document vectors. We abstract Type III to a general framework:

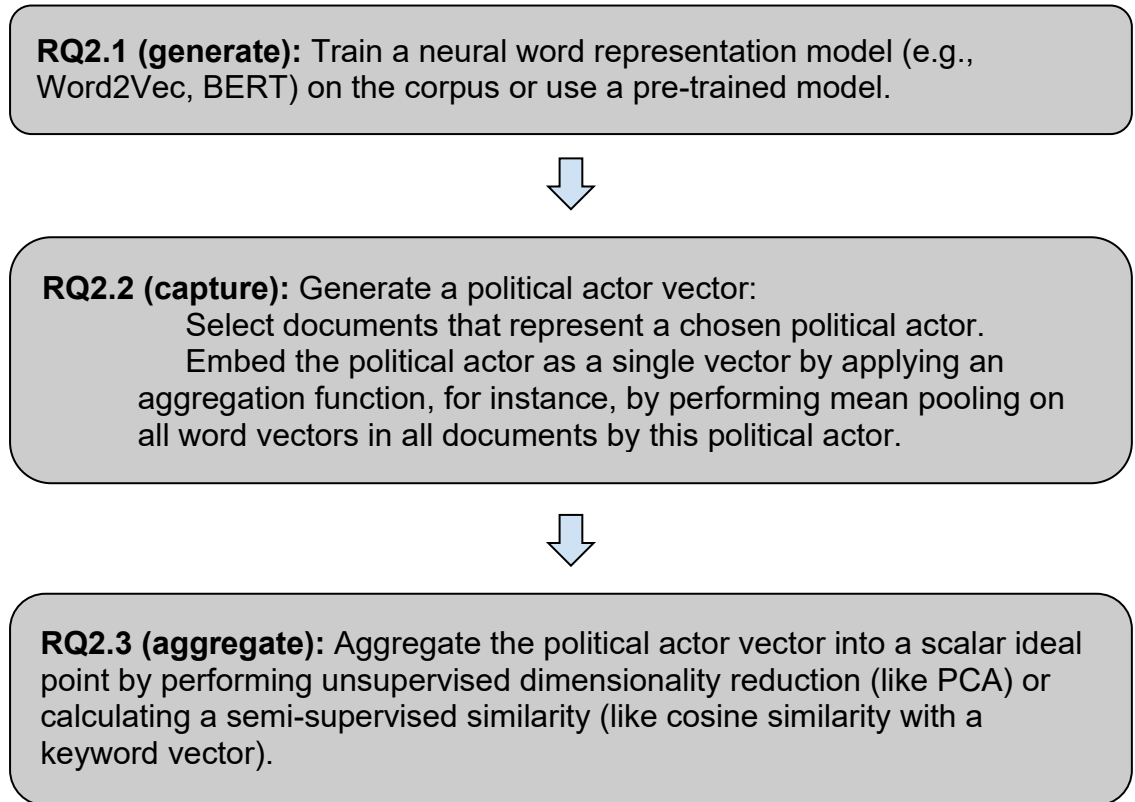

**RQ2.1 (generate):** Train a neural word representation model (e.g., Word2Vec, BERT) on the corpus or use a pre-trained model.

**RQ2.2 (capture):** Generate a political actor vector:
Select documents that represent a chosen political actor.
Embed the political actor as a single vector by applying an aggregation function, for instance, by performing mean pooling on all word vectors in all documents by this political actor.

**RQ2.3 (aggregate):** Aggregate the political actor vector into a scalar ideal point by performing unsupervised dimensionality reduction (like PCA) or calculating a semi-supervised similarity (like cosine similarity with a keyword vector).

Figure 5: Type III framework.

Nanni et al. (2022) propose *SemScale* and *SemScore*. While the former is unsupervised and the latter semi-supervised, both capture variance by computing pairwise similarities between document embeddings and constructing a similarity graph. They aggregate variance by propagating positions across this graph, using graph-based ranking algorithms such as PageRank.

While implementation details vary, these algorithms share a core logic: they rely on variance in semantic similarity in embedding spaces for measuring ideal point constructs. This approach abstracts from surface-level word use and focuses on the co-occurrence of words encoded in neural vector representations.

### 5.2.4 Type IV: LLM-based Algorithms

This type includes all approaches that use generative LLMs for prompt-based CT-IPE. LLMs act as large-scale knowledge bases and rely on the same distributional semantic logic as word embeddings (Type III) but at a grander scale and with prompt-based accessibility. It is important to note that in Type IV, the generate variance step is always performed by pre-trained large language models at the time of this writing. Most researchers cannot influence this step directly, as the models are trained externally. However, researchers can influence which LLM they choose for a research question. Type IV uses prompts to access specific subspaces of an LLM's embedding space to capture relevant variance. A simplified prompt might ask the model to "act as a helpful ideal point estimator for a left-right axis."

For instance, Le Mens and Gallego (2025) capture variance by prompting an LLM to place political texts on a focal dimension. They aggregate variance by averaging the model's responses across texts to estimate the ideal points of U.S. Senators. The Type IV framework follows three steps:

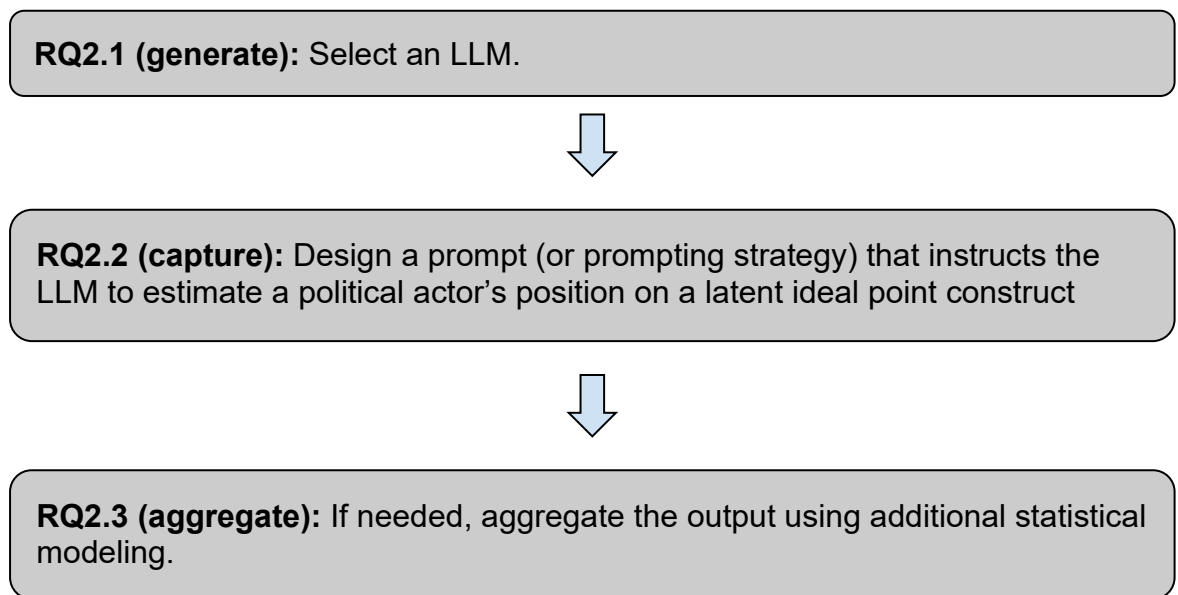

Figure 6: Type IV framework.

As Figure 6 illustrates, modeling decisions of Type IV are abstracted more than in earlier approaches. While this need not reduce validity, it requires heightened transparency and methodological care.

One other algorithm paper from Type IV by O'Hagan and Schein (2024) addresses this concern, offering a detailed reflection on validity and experimenting with techniques such as chain-of-thought (CoT) prompting to guide LLM behavior. Burnham (2024) introduces *Semantic Scaling*, prompting LLMs to produce survey-like data in the variance capture step that is then scaled using item response theory (IRT) in the aggregation step.

Wu et al. (2023) propose *Language Model Pairwise Comparison* (LaMP), where an LLM compares lawmakers in pairs rather than estimating positions directly. In a follow-up paper, Wu et al. (2024) extend the method through *Concept-Guided Chain-of-Thought (CGCoT)*. This strategy combines structured prompting with interpretability, mirroring the guidance also often given to humans in manual annotation processes.

In summary, algorithms of Type IV leverage the semantic and conversational power of LLMs but shift variance modeling into the prompt design, bringing new opportunities and challenges. For instance, point estimates of many IPE studies are likely in the training data of many LLMs. Therefore, the question with LLMs is whether they are new techniques of CT-IPE or rather databases for retrieving summarizations of other algorithms (see Balloccu et al. (2024) on the problem of data leaking). Overall, LLMs introduce new concerns of algorithmic model bias but are flexible and low-friction tools for CT-IPE.

### 5.2.5 Directly Comparing the Four Types

The typology is designed not merely to categorize but to enable critical comparison: by abstracting algorithms into types, we make their methodological differences explicit. The typology matrix in Figure 7 summarizes all four types and how they approach the problem of modeling variance.

| | Word-frequency (Type I) | Topic modeling (Type II) | Word embedding (Type III) | LLMs (Type IV) |
|---|---|---|---|---|
| **Generate Variance** | Word frequencies per document | Topic distributions per document (e.g. LDA, poisson factorization) | Word or document embeddings (e.g. FastText, BERT) | LLM-internal word embeddings (ChatGPT, LLaMA) |
| **Capture Variance** | Find relevant variance in frequencies (e.g. weighting, reference texts) | Find relevant variance in topic distributions (e.g. weighting, topic shifts) | Find relevant variance in embeddings (e.g. keywords, similarity networks) | Find relevant variance in LLM embeddings (e.g. CoT, pairwise comparison) |
| **Aggregate Variance** | Aggregate relevant word frequency variance (e.g. MLE, clustering) | Aggregate relevant topic variance (e.g. VI, MCMC) | Aggregate relevant embedding variance (e.g. PCA, cosine similarity) | Aggregate relevant LLM variance (e.g. IRT, averaging) |

Figure 7: Typology matrix. Abbreviations: LDA: Latent Dirichlet Allocation; CoT: Chain of Thought prompting; MLE: Maximum Likelihood Estimation; VI: Variational Inference; IRT: Item Response Theory; Markov Chain Monte Carlo (MCMC).

For instance, one can see from the matrix that Type I and Type II differ in what is being aggregated and how. Likewise, the matrix shows that while Type III captures variance through similarity in embedding spaces, Type IV relies on comparable data structures but uses a different mode of accessing them, namely prompting. This shifts the modeling decision in the capture step from statistical vector analysis to prompt design.

Therefore, the matrix shows that the concept of relevant variance for the estimation of an ideal point differs by type. We refer here to what we defined as relevant variance in section 3. Accordingly, how algorithms aggregate such captured relevant variance differs across the four types.

By combining the two RQs presented so far, namely development context and methodological structure, our review provides a deeper synthesis than either perspective alone. The development context variables show how algorithms were designed, while the variance framework shows how they function as measurement strategies. This integration prepares the subsequent section, where we translate our review into practical guidance for applied research

## 5.3 Guiding Algorithm Choice

The aim of this section is not to identify a single "best" algorithm but to clarify the trade-offs and contextual dependencies that shape choices. Three dimensions matter most: (1) theoretical alignment, (2) data, compute, and transparency, and (3) validation. These dimensions capture how well a method reflects the construct of interest, whether it can be implemented under real-world constraints, and how convincingly its results can be justified.

### 5.3.1 Social Science Theory

Algorithm choice must begin with theory. CT-IPE is not only a technical procedure but an attempt to measure latent constructs defined by theoretical expectations. As discussed in section 5.1.1, reviewed algorithms are grounded in diverse theoretical perspectives.

Therefore, the key task for applied research is alignment: the chosen algorithm should reflect the construct of interest, the country context, and the textual domain under study (see Figure 8). For example, when the research question revolves around the issue of European integration in German parliamentary speeches, an algorithm developed on this setup will be a good starting point. When party competition is theorized spatially, embedding-based methods (Type III) may fit well because of their natural spatial setup. Likewise, Type II algorithms are often developed with framing as a theory in mind. Conversely, when theoretical expectations hinge on the salience of specific words, word-frequency methods (Type I) offer an operationalization. Type IV allows flexible prompts that can mirror different theoretical constructs, albeit raising other trade-offs (see 5.3.2 and 5.3.3).

### 5.3.2 Data and Computational Power

The types also differ in technical properties that matter for applied work (see Figure 8). Counting words (Type I) to generate variance produces fewer parameters and thus demands less data and compute than estimating large parameter spaces, such as those based on word co-occurrence matrices (embeddings, Type III). Accordingly, the types scale ordinally in computational and data demand (Type I < II < III < IV). Both factors inversely relate to transparency, where code availability also comes into play. Researchers with limited computing power and data who seek easily traceable models may gravitate toward Type I or II, since one cannot reliably estimate more parameters than the available data support, and an increased number of parameters raises computational cost. In contrast, those who want to leverage 'world knowledge' from large datasets and have the compute available to digest them, or who need the flexibility of prompts, may consider Type III or IV, while being mindful of their transparency trade-offs (see next section).

### 5.3.3 Validation, Robustness, and Transparency

Validation is not merely a post-estimation step but a factor that should inform algorithm choice. Algorithms vary in how their results can be validated and in what robustness checks are available.

Across all types, the cornerstone of validation is comparison against gold standards (see section 5.1.3). However, how transparently researchers can interpret divergences between algorithm estimation and the gold standard differs across types and algorithms. In contrast to validity, robustness checks or tests for overfitting (e.g., bootstrapping) vary more by algorithm and can place very different demands on applied researchers.

For Type I, validation is comparatively straightforward because the link between word counts and estimated positions is relatively transparent and based on relatively simple calculations. This makes it easier to trace why a text is placed at a given position and to check plausibility against external benchmarks. Robustness can be supported through methods like Monte Carlo simulation (Lo et al. 2014) or bootstrapping (Jentsch et al. 2020).

For Type II, validation is less straightforward because estimated positions depend on latent topics that must first be interpreted. Applied researchers must judge whether these topics plausibly represent the theoretical construct before validation against benchmarks becomes meaningful. Robustness strategies in this family include semantic coherence measures (Diaf & Fritsche 2022) and downstream tasks such as classification to test topics' stability (van der Zwaan et al. 2016).

For Type III, validation is at least as demanding as for Type II. Applied researchers must interpret scatterplots of embeddings and similarity measures across many terms and hyperparameters. This complexity makes it easy to lose track of how estimates are generated. While external validation benchmarks remain central, robustness checks are again essential to establish estimation credibility. Strategies include resampling across training runs (Nanni et al. 2022) and bootstrapping (Rasmussen et al. 2023). Such checks are essential when leveraging the richer semantic information that embeddings provide.

For Type IV, validation is the most challenging because results are generated by highly opaque, large pretrained models, often within proprietary software. Applied researchers face the difficulty that neither the training data nor the full parameterization is (easily) accessible, making it virtually impossible to transparently trace how outputs are produced.  Even when training data is disclosed, as in some open models, its sheer scale and heterogeneity prevent researchers from linking the composition of the data to specific ideal point estimates in any systematic way. These factors are often reffered to as the 'back-box' nature of LLMs. Validation strategies for Type IV are therefore crucial but still developing. While external gold standards remain indispensable (with the above mentioned caveats of interpretation), additional interpretability tools such as structured prompting or CoT reasoning (Wu et al. 2024) have been developed. Some studies test robustness via internal model consistency, for example, by generating policy texts for given ideal points and re-scoring them against anchors (O'Hagan & Schein, 2024). These procedures are resource-intensive and far less standardized than for other types.

In sum, validation and robustness demands rise from Type I to Type IV: earlier algorithms (see the timline in Figure 2) permit established checks and are more transparent, whereas newer and more complex ones require elaborate or untested strategies. Overall, researchers should balance theoretical integration, technical properties, and validation possibilities when selecting an algorithm.

| | Type | Data size | Transparency | Compute | Text domain | Country domain | Construct | Code available |
|---|---|---|---|---|---|---|---|---|
| Laver et al. (2003) | Type I | Low | High | Mid | Manifestos | Britain, Ireland, Germany | Left-right, economic, social | green |
| Martin & Vanberg (2007) | | | | | Manifestos | Britain | Left-right, economic | green |
| Slapin & Proksch (2008) | | | | | Manifestos | Germany | Left-right, economic, societal, foreign policy | green |
| Lo et al. (2014) | | | | | Manifestos | Germany, Ireland, Netherlands, Sweden | Left-right | green |
| Hemphill et al. (2016) | | | | | Social media posts | United States | Political polarization | green |
| Lauderdale & Herzog (2016) | | | | | Manifestos | Ireland, United States | Left-right, government-opposition | green |
| Temporao et al. (2018) | | | | | Social media posts | Canada, New Zealand | Left-right | green |
| Gentzkow et al. (2019) | | | | | Speeches | United States | Partisanship | green |
| Jentsch et al. (2020) | | | | | Manifestos | Germany | Left-right | green |
| Diaf (2023) | | | | | Manifestos | Germany | Left-right | red |
| van der Zwaan et al. (2016) | Type II | Low | Mid | Mid | Manifestos | Netherlands | Left-right, topic-specific positions | green |
| Vafa et al. (2020) | | | | | Speeches, social media posts | United States | Liberal-conservative | green |
| Kaneko et al., (2021) | | | | | News articles | Japan | Left-right | green |
| Diaf & Fritsche (2022) | | | | | Manifestos | Germany | Partisanship | red |
| Hofmarcher et al. (2025) | | | | | Speeches | United States | Partisanship | green |
| Gordon et al. (2020) | Type III | Low | Mid | Mid | Social media posts | United States | Left-right, authoritarian-libertarian | red |
| Rheault & Cochrane (2020) | | | | | Speeches | Britain, Canada, United States | Left-right | green |
| Nanni et al. (2022) | | | | | Speeches, manifestos | European Union | Left-right, european integration | green |
| H. R. Rasmussen et al. (2024) | | | | | Social media posts | United States, Denmark, Sweden, Germany, Italy | Online political hostility | green |
| Maurer et al. (2024) | | | | | Social media posts | Germany | Policy positions | green |
| Wu et al. (2023) | Type IV | High | Mid | High | - | United States | Liberal-conservative, gun control, abortion | red |
| Wu et al. (2024) | | | | | Social media posts | United States | Aversion to parties | green |
| O'Hagan & Schein (2024) | | | | | Social media posts | United States | Left-right | red |
| Burnham (2024) | | | | | Social media posts | United States | Liberal-conservative | green |
| Le Mens & Gallego (2025) | | | | | Manifestos, speeches, social media posts | United States, Britain, European Union | Left-right, economic, social | green |

Legend: ■ Low ■ Mid ■ High

Figure 8: Algorithm choice guide. Reading example: High data/compute: "The algorithm requires a high amount of compute/data for training and inference combined."

# 6 Limitations

We are aware of some limitations of our paper that we wish to discuss here. By focusing on high-level variance modeling, we abstract from specific implementation choices that can significantly affect results in practice. While this abstraction is necessary to enable synthesis across such a diverse field, it risks overlooking meaningful methodological nuances critical for methods and applied research.

However, we found that using variance as a conceptual tool rather than digging into nuances enables comparing the algorithms at the high level we intended, which is appropriate for a literature review. Digging deeper will require a methodological paper of its own since the algorithms are too different in terms of statistical modeling details. Additionally, one needs data from a computational experiment to investigate this direction more profoundly and compare the empirical variance between the algorithm estimations. We return to this point in the discussion section, where we outline a benchmarking experiment.

Another point is that some algorithms that some readers might expect here did not make it into the review. One of these is *Latent Semantic Scaling* by Watanabe (2021). It is conceptually very close to Type III, but does not frame itself around ideal point estimation in the sense of our review. As mentioned earlier, the theoretical possibility of an algorithm being suited to CT-IPE is not a sufficient condition for inclusion, as this would necessarily entail a deeper methodological evaluation. The *Class Affinity Model* is another algorithm that did not make it and is even implemented in some software packages (Perry and Benoit 2017). To our knowledge, no peer-reviewed version exists, and we could therefore not include it. We refer the reader to online Appendices 1, 3, and 5, where we trace which papers were included at which step and why.

# 7 Discussion and Outlook

Our results map out a fragmented field of research and provide guidance for algorithm choice in applied research. Beyond this immediate contribution, we also want to highlight some open questions and future directions for the field. In what follows, we discuss two areas where further work is needed: (1) connecting CT-IPE more systematically to theoretical ideas, (2) conducting a benchmarking experiment.

## 7.1 Theory: Integrating Broader Theoretical Foundations

Estimating a political actor's "ideal" position requires more than analyzing what they say; it also involves considering intentions, commitments, and strategic behavior. Mandate theory offers one way forward: Lemmer's (2023) *Specificity Space Model* integrates mandate theory with IPE by accounting for the specificity of political statements (see also Subramanian et al. 2019). Related perspectives from psychological positioning theory and distributional semantics provide further opportunities to enrich models of how actors construct and communicate political positions (Van Langenhove 2020; Mickus 2022).

CT-IPE can also benefit from engaging more deeply with the non-computational IPE literature, which has long debated and refined manual text-based approaches (Budge 2001; Lowe et al. 2011; Volkens et al. 2013; Dolezal et al. 2016; Lemmer 2023). Integrating these theoretical perspectives more systematically into CT-IPE would strengthen construct definition and operationalization, helping to align computational strategies with the complex realities of political positioning.

## 7.2 Benchmarking: Building Systematic Evaluation Standards

Validation practices in CT-IPE are diverse, ranging from comparisons with expert surveys to face-validity checks, but they remain fragmented and inconsistent across studies. While individual contributions have introduced valuable strategies, the field lacks systematic benchmarking. Although there are experimental comparisons of algorithms (e.g., Hjorth et al.

2015; Koljonen et al. 2022), we argue that these should be undertaken more systematically, providing "experimental, substantive, and statistical evidence" (Grimmer and Stewart 2013, 271).

 A critical next step is a large-scale benchmark experiment: applying a broad set of CT-IPE algorithms to a shared, diverse dataset. Similar efforts have proven valuable in tasks like sentiment analysis (Ribeiro et al. 2016) or topic modeling (Doan and Hoang 2021). Such benchmarking would move the field beyond isolated comparisons and enable empirically grounded methodological choices.

We therefore call for a shift from merely replicating results to empirically examining how different algorithms model variance. Treating divergence in outcomes—not only convergence—as the dependent variable would help identify which methods yield novel perspectives and under what contexts these emerge.

## 7.2.1 Metrics for Benchmarking Across Variance Steps

A large-scale benchmarking experiment should not only compare final point estimates but also evaluate how algorithms perform at each step of the variance framework—generation, capture, and aggregation. At the generation stage, metrics could include the number and type of features extracted (e.g., vocabulary size, topic counts, embedding dimensions) and the variance distribution across them. At the capture stage, one might assess how much variance is retained or discarded, for example, by measuring the explained variance of principal components or the coherence of topics. At the aggregation stage, comparisons could involve the dispersion of actor-level estimates, the sensitivity of results to document sampling, or the extent to which aggregation choices introduce bias or dampen variance. Variables like runtime or memory usage are easy to track and would provide valuable information for researchers.

## 7.2.2 Challenges of Developing Overarching Metrics

Designing such metrics poses significant challenges. The three variance steps are implemented differently across algorithmic types and single algorithms, making it difficult to define commensurable indicators. For instance, "variance explained" in a PCA-based embedding model does not directly map onto coherence scores in a topic model. Moreover, preprocessing, dimensionality reduction, or anchoring strategies can influence results, making it unclear whether differences reflect algorithmic logic or design decisions. Thus, any benchmark must strike a balance between comparability and respecting the internal logic of each method. We therefore envision a benchmarking framework that combines step-specific diagnostics with outcome-oriented metrics—such as comparison with gold standards—allowing researchers to disentangle where and why algorithms diverge.

Over the past two decades, CT-IPE has largely advanced by following successive NLP trends—moving from word counts to topics, embeddings, and most recently LLMs. Yet further proliferation risks deepening fragmentation. The field may now benefit more from synthesizing and systematically comparing what already exists. As Lowe (2016, 2) wryly noted, "There should be one – and preferably only one – obvious way to do it." While not meant literally, the remark underscores the need for greater integration and shared standards rather than the pursuit of ever-new methods.

# Acknowledgments

The authors gratefully acknowledge the invaluable feedback and support provided by Mario Haim, Barbara Plank, Valerie Hase, and Verena Schwald. We also thank the discussant (Petro Tolochko), chair (Marc Ratkovic), and audience of the panel "Innovations in Ideological Estimation and Latent Modeling" for their feedback at the 2025 annual COMPTEXT conference in Vienna.

# Competing Interests

The authors declare none.

# Funding statement

The first author received financial support for this article via the Graduate Center for Doctoral Researchers at the Bavarian Research Institute for Digital Transformation. Financial sponsors played no role in the design, execution, analysis and interpretation of data, or writing of the study.

# Data availability

Datasets with all retrieved database records for the literature review can be found in the data folder in the following OSF repository (URL anonymized for review):

https://osf.io/xpkaf/?view_only=5a6ac2786b664c3591e2ac34ff0f1cc9

# Supplementary material

Supplemental material, replication data and code for this article is available online and via OSF (URL anonymized for review):

https://osf.io/xpkaf/?view_only=5a6ac2786b664c3591e2ac34ff0f1cc9